\pdfoutput=1

\documentclass[11pt]{article}

\usepackage[]{EACL2023}

\usepackage{times}
\usepackage{latexsym}
\usepackage{graphicx}

\usepackage[T1]{fontenc}

\usepackage[utf8]{inputenc}

\usepackage{multirow}
\usepackage{amsmath}
\usepackage{algorithm}
\usepackage{algorithmic}
\usepackage{latexsym}

\usepackage{inconsolata}

\title{Find Parent then Label Children: A Two-stage Taxonomy Completion Method with Pre-trained Language Model}

\author{Fei Xia$^{1,2}$\thanks{$^{*}$ These authors contribute this work equally.}, Yixuan Weng$^{1*}$, Shizhu He$^{1,2}$, Kang Liu$^{1,2}$ and Jun Zhao$^{1,2}$ \\
	$^1$ National Laboratory of Pattern Recognition,, Institute of Automation, CAS \\
	$^2$ School of Artificial Intelligence, University of Chinese Academy of Sciences\\
\texttt{xiafei2020@ia.ac.cn,wengsyx@gmail.com}\\
\texttt{\{shizhu.he, kliu, jzhao\}@nlpr.ia.ac.cn}
}

\begin{document}
\maketitle
\begin{abstract}
Taxonomies, which organize domain concepts into hierarchical structures, are crucial for building knowledge systems and downstream applications. As domain knowledge evolves, taxonomies need to be continuously updated to include new concepts.
Previous approaches have mainly focused on adding concepts to the leaf nodes of the existing hierarchical tree, which does not fully utilize the taxonomy's knowledge and is unable to update the original taxonomy structure (usually involving non-leaf nodes). In this paper, we propose a two-stage method called ATTEMPT for taxonomy completion. Our method inserts new concepts into the correct position by finding a parent node and labeling child nodes. Specifically, by combining local nodes with prompts to generate natural sentences, we take advantage of pre-trained language models for hypernym/hyponymy recognition. Experimental results on two public datasets (including six domains) show that ATTEMPT performs best on both taxonomy completion and extension tasks, surpassing existing methods.
\end{abstract}

\section{Introduction}

Taxonomies\footnote{In this paper, we mainly focus on the taxonomy represented as tree rather than directed acyclic graph, because trees are the mainstream form at present, such as the online catalog taxonomies of Amazon and Yelp. } are an important form of domain knowledge that organize concepts into hierarchical structures, representing ``hypernym-hyponym” relationships among concepts in the form of trees or directed acyclic graphs  \cite{10.1145/3366423.3380132}. Taxonomies are essential components of knowledge systems such as ontologies and knowledge graph~\cite{10.1145/3394486.3403145}, and are widely used in various downstream applications, including search engineering~\cite{yin2010building}, recommendation systems \cite{JinHuang2019TaxonomyAwareMR,zhang2014taxonomy}, and information filtering \cite{demeester2016lifted}.

\begin{figure}[t]
	\centering
	\includegraphics[scale=0.625]{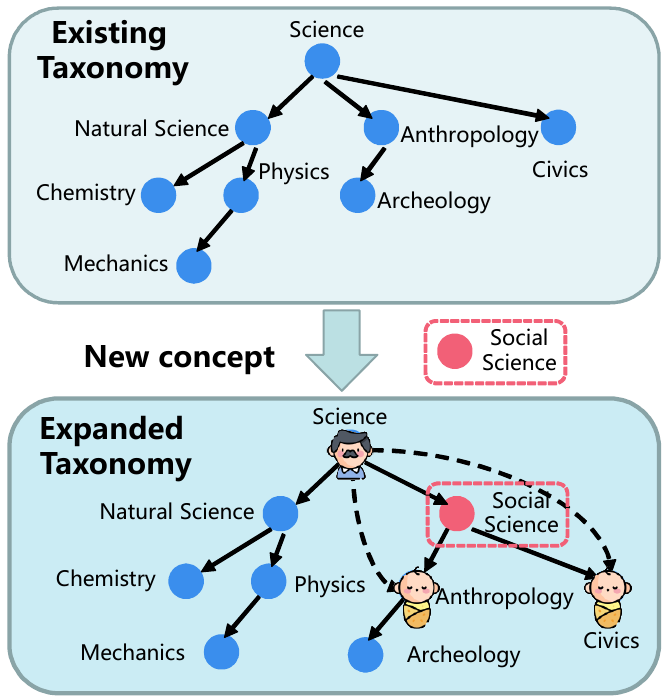}
	\caption{An example of taxonomy completion for a non-leaf node. The new concept ``Social Science" needs to be inserted into the correct position in the existing taxonomy.}
	\label{fig1}
		\vspace{-0.5cm}
\end{figure}

As domain knowledge continues to evolve, especially with the rapid growth of web content, new concepts are constantly emerging. In order to stay current, original taxonomies must incorporate these new concepts and adapt their hierarchical relationships. For example, as shown in Figure \ref{fig1}, with the advancement of sociology and science, the concept of "Social Science" should be added to the science knowledge system, and the original structure should be adjusted accordingly.

However, existing taxonomies are primarily constructed by human experts \cite{JiamingShen2018HiExpanTT}. Manual extraction of domain concepts and detection of hierarchical relationships by domain experts is both time-consuming and labor-intensive, and may result in missing important concepts and relationships.

To extend existing taxonomies automatically, researchers have proposed the tasks of taxonomy expansion (TE) and taxonomy completion (TC). Both tasks aim to append new nodes (concepts) to a given taxonomy . The main difference is that TE focuses on identifying the parent of a given node (usually a leaf node), while TC aims to identify both the parent and child nodes. As illustrated in Figure \ref{fig1}, TE would aim to identify the parent node of  ``Social Science”, while TC would also aim to identify the child nodes of  ``Social Science”.

Recently, researchers have been focusing on using pre-trained language models, such as BERT \cite{devlin2018bert}, to improve the performance of taxonomy expansion~\cite{liu-etal-2021-temp, takeoka-etal-2021-low}. For example, TEMP~\cite{liu-etal-2021-temp} appends new concepts to leaf nodes and generates candidate taxonomy paths, then uses a pre-trained model for ranking and selecting the best path. Musubu~\cite{takeoka-etal-2021-low} generates candidate ``hypernym”-``new concept” pairs using Hearst patterns~\cite{MartiAHearst1992AutomaticAO}, and relies on pre-trained knowledge to identify the optimal hypernym node for the new concept. These proposed models have greatly improved the effectiveness of taxonomy updates, thanks to the improved generalization performance of pre-trained language models \cite{liu-etal-2021-temp}.

Although current TE\&TC methods have achieved good results, there are several main issues that need to be addressed. Firstly, existing TE methods struggle to extend non-leaf nodes or perform poorly in this task \cite{zhang2021tmn}. Secondly, while existing TC methods can extend both leaf and non-leaf nodes, they may be less effective in leaf node expansion than specialized TE methods \cite{liu-etal-2021-temp}, potentially due to a lack of sufficient utilization of knowledge. Furthermore, these methods often require large amounts of labeled samples or external resources, which are not always available \cite{takeoka-etal-2021-low}. Lastly, current TC methods do not typically involve modifying the nodes of the original taxonomy system (all original parent-child relationships are preserved after adding nodes to the taxonomy). However, the insertion of new nodes can modify the relationship of the original nodes. For example, the insertion of ``Social Science" in Figure \ref{fig1} would change the relationship between ``Science-Anthropology" from father-son to grandfather-grandson.

To address these issues, we propose \textbf{A} \textbf{T}wo-stage \textbf{T}axonomy compl\textbf{E}tion \textbf{M}ethod with \textbf{P}re-\textbf{T}rained Language model (ATTEMPT), which inserts new concepts into the correct position by identifying a parent node and labeling child nodes.

In the first stage of our proposed method, we use the ``Taxonomy-path Prompt with Pre-trained model" (PPT) approach to take advantage of the local information of the taxonomy path and convert it into natural language using a prompt method, which helps to better utilize the implicit knowledge of the pre-trained model. Additionally, the pre-trained model's extensive knowledge reserve allows us to avoid the need for external resources and large amounts of labeled data. In the second stage, we propose the ``Multiple Nodes Labeling" (MNL) method, which jointly identifies each child node and better utilizes the interdependence between nodes, resulting in more accurate node type prediction (including father-son, sibling and other relationships). Additionally, MNL allows for modification of the original taxonomy nodes and simultaneous annotation of multiple child nodes.

We conduct detailed experiments on two public datasets (including six domains) to evaluate the effectiveness of our proposed method, ATTEMPT, in leaf and non-leaf node expansion. Specifically, for leaf nodes, our parent-finding method (PPT) outperforms the best baseline by 8.2\% in accuracy. For non-leaf nodes, our children-finding method (MNL) improves by 21\% and 20.3\% respectively in accuracy and average F1 score, compared to a pair-wise classification method. On the overall task, our proposed method (ATTEMPT) outperforms other methods by 2.4\% in average F1 score.

In summary, the main contributions of this paper include:

$\bullet$ The proposal of a two-stage taxonomy expansion method, ATTEMPT, that inserts new concepts into the correct position by identifying a parent node and labeling child nodes.

$\bullet$ The introduction of a multiple-nodes labeling method, MNL, for the children-finding stage, which allows for the label of zero to multiple children nodes of a given node simultaneously and modification of the original taxonomy nodes.

$\bullet$ The demonstration of the effectiveness of our approach through experiments on two public datasets (including six domains), with the best performance obtained in both non-leaf and leaf node expansion.

\section{Related Work}
Taxonomy construction aims to build a tree-structured taxonomy with a set of terms from scratch. Existing methods can be roughly divided into two categories. The first is an unsupervised method to construct the taxonomy based on clustering  \cite{alfarone2015unsupervised,zhang2018taxogen,shang2020nettaxo}. The terms are grouped into a hierarchy based on hierarchical clustering or topic models \cite{downey2015efficient}. Each node of this taxonomy is a collection of topic-indicative terms, different from the taxonomy in this paper (each node represented by one individual term). The other approach constructs a taxonomy based on terms, where each node represents a term concept \cite{cocos2018comparing,dash2020hypernym}. Hypernymy detection models are often used for this task. For example, pattern-based \cite{agichtein2000snowball,jiang2017metapad,roller2018hearst} or distributional models \cite{yin2018term,wang2019family,dash2020hypernym} extract the hypernymy for a given query node and then organize them into a tree structure. 

Creating a taxonomy from scratch is labor-intensive. In many scenarios, such as e-commerce, some taxonomies may already be deployed in online systems, which involves the demand   of taxonomy extension. QASSIT \cite{cleuziou-moreno-2016-qassit} is a semi-supervised vocabulary classification method, mainly based on genetic algorithms.  The TAXI \cite{panchenko-etal-2016-taxi} system uses a taxonomy induction method based on lexico-syntactic patterns, substrings, and focused crawling. Later, TaxoGen \cite{ChaoZhang2018TaxoGenUT} uses term embeddings and hierarchical clustering to construct topic taxonomies recursively. TEMP \cite{liu-etal-2021-temp} is a self-supervised classification extension method that trains models with a new dynamic margin loss margin function.

Taxonomy completion \cite{zhang2021tmn} is a recently proposed task that aims to find appropriate hypernyms-hyponyms for new nodes, not just hypernyms. GenTaxo \cite{zeng2021enhancing} gathers information from complex local structural information and learns to generate full names of concepts from corpora. TMN \cite{zhang2021tmn} focuses on channel gating mechanisms and triplet matching networks. CoRel relies on concept learning and relation transferring to build a seed-oriented topic taxonomy. 

But the above mentioned methods also have some issues. The addition of new nodes may also lead to changes in the original taxonomy. The taxonomy completion task only finds the hyponyms of a given node, which cannot modify of the original taxonomy. GenTaxo \cite{zeng2021enhancing} requires a large amount of training data to learn enough information, and CoRel \cite{huang2020corel} focuses more on topic taxonomy than the taxonomy of individual terms. Other works such as CGExpan \cite{zhang2020empower} use the automatically generated class names and the class-guided entity selection module for entity expansion. However, CGExpan \cite{zhang2020empower} is more on the entity set than the tree taxonomy.

In addition, although the above methods can find both hypernyms and hyponyms of a given query node, they do not make sufficient use of the pre-trained model or do not use the pre-trained model at all \cite{zhang2021tmn,zeng2021enhancing}. This may lead them to perform poorly on the hypernym recognition task, inferior to the specialized taxonomy extension methods of the pre-trained model \cite{liu-etal-2021-temp}. And most methods of taxonomy extension cannot perform well on the task of taxonomy completion \cite{zhang2021tmn}. We are dedicated to finding an approach that works in both tasks.

\begin{figure*}[t]
	\centering
	\includegraphics[scale=0.637]{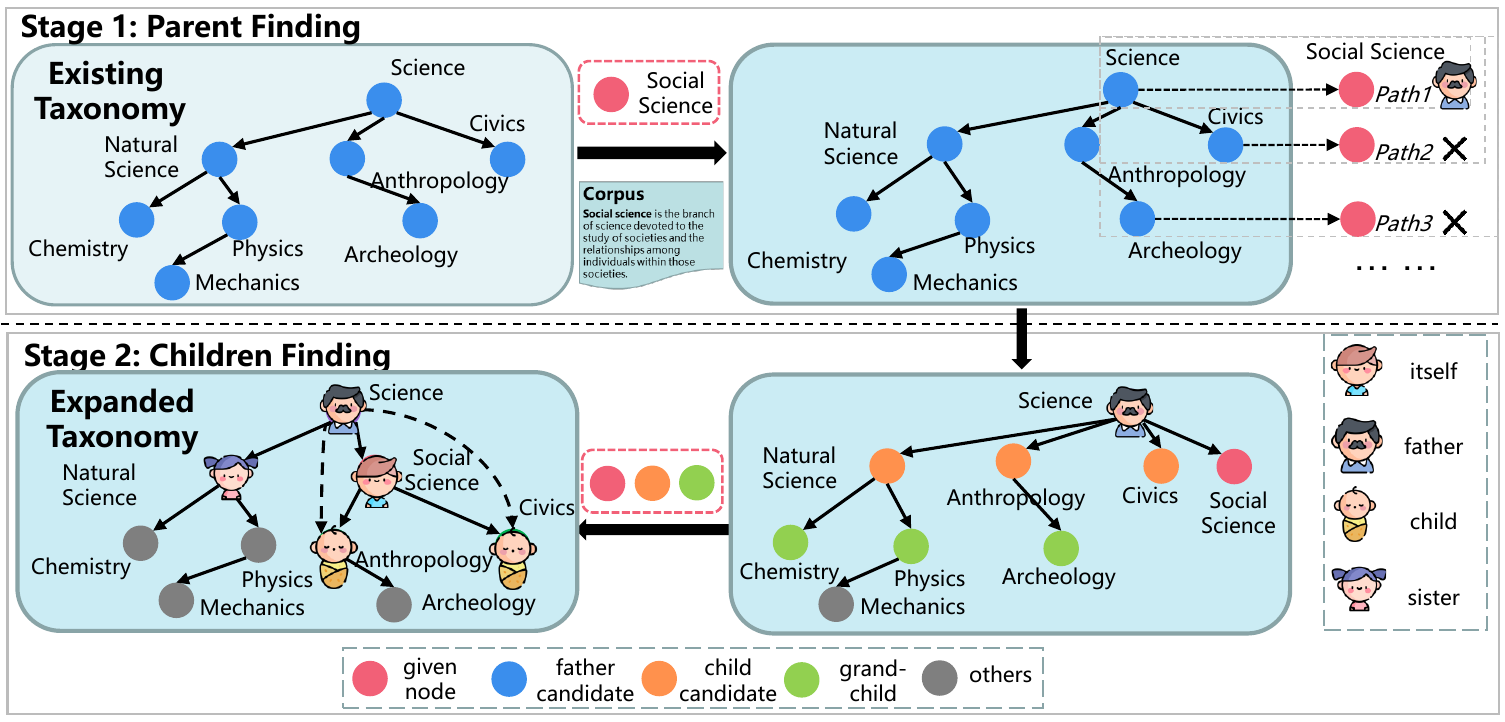}
	\caption{An overview of the proposed method ATTEMPT. ATTEMPT consists of two stages. Given an existing taxonomy and the new concept term (``Social Science"), the first stage is to find the correct parent ("Science") and the second stage is to find all possible children ("Anthropology" and "Civics").}
	\label{fig2}
\end{figure*}

\section{Method}
Given an existing taxonomy $T=(V, E)$ and a set of new terms $V^{\prime}$, where $V$ is a set of terms, and $E$ is a set of " hyponym- hypernym" relationships between terms, the task of Taxonomy completion is to insert the new terms $v^{\prime} \in V^{\prime}$ into the appropriate position of the existing taxonomy $T$ one by one and extend them into a more complete taxonomy $\tilde{T}=(\tilde{V}, \tilde{E})$. 

Figure \ref{fig2} provided illustrates the overall structure of the ATTEMPT method, which is broken down into two main stages: the parent finding stage and the children finding stage. These two stages work together to identify the relationships between terms in the taxonomy, specifically determining the parent and children of a given term.

\subsection{Stage one: Parent Finding}
The first stage of the process is to identify the parent node of a given node in the taxonomy. For example, finding the parent node ``science" for the node ``social science" in Figure \ref{fig2}.

\subsubsection{TEMP}
The TEMP method~\cite{liu-etal-2021-temp} is the first approach to use pre-trained contextual encoders as the core component for taxonomy extension. The pre-trained contextual embeddings are useful for capturing relationships between terms because they have been trained on a large corpus. TEMP predicts the location of new concepts by ranking the generated taxonomy paths. A taxonomy path of a new term (ND) in the tree-structured taxonomy is the unique path from that term to the root of the taxonomy. The taxonomy path is represented as $P = [ROOT, N_1, N_2, ..., N_D]$, where $D$ is the depth of the ND and $ROOT$ is the root of the taxonomy. In the taxonomy, $N_{i-1}$ is the parent of $N_i$. TEMP generates taxonomy paths for each term, then adds the new term to be expanded to the end of each path to form new paths. Finally, the new paths are ranked and the highest-scoring path is chosen as the correct parent term.

Equation \ref{eq1} describes how TEMP uses a contextual encoder to return a sequence of vectors, given a term's definition $S$ and an arbitrary taxonomy path $P$.

{\small
\begin{equation}
\operatorname{Encoder}(S, P)=v_{[\mathrm{CLS}]}, v_{1}, \ldots, v_{[\mathrm{SEP}]}, v_{p_{d}}, \ldots, v_{\text {root }}
\label{eq1}
\end{equation}
 }
The TEMP method, which uses pre-trained contextual encoders to model taxonomy paths, has been an inspiration for our work. However, TEMP also has some limitations. One of the main limitations is that it can only expand new leaf nodes. Additionally, TEMP has some issues such as:

1) Limited use of local information - although TEMP uses paths to narrow the search range within the taxonomy tree, the problem of too long paths can still arise. In such cases, distant relationships may have a limited impact on the determination of leaf nodes.

2) Inadequate utilization of pre-trained model - TEMP only connects the nodes of the path using special tokens such as $[SEP]$ or $[UNK]$, which does not fully leverage the knowledge encoded by the pre-trained language model.


 \begin{figure*}[t]
	\centering

	\includegraphics[scale=0.633]{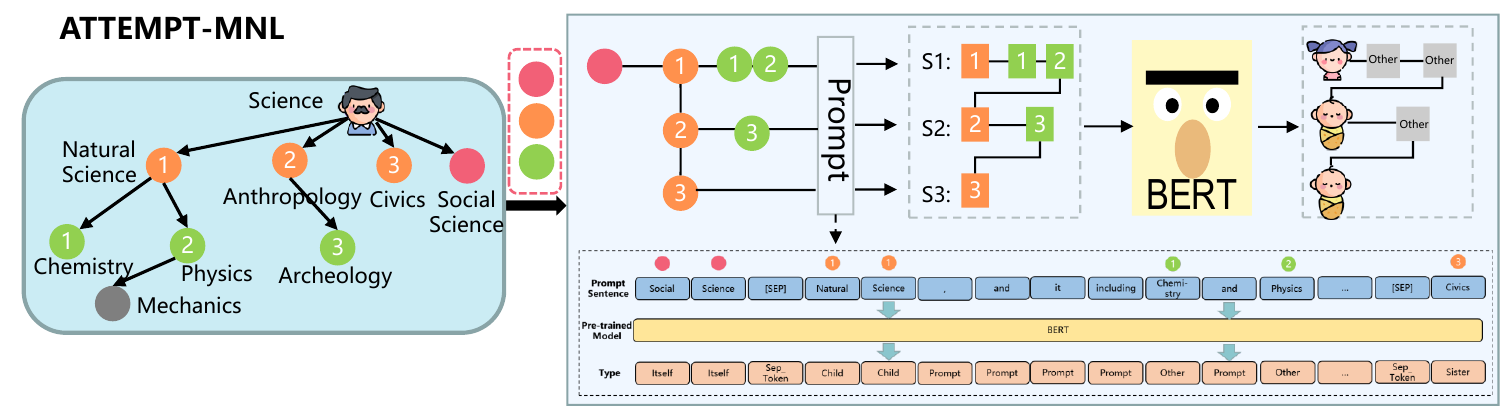}

	\caption{An overview of the MNL (multiple nodes labeling). We concatenate the current node (red), parent node, candidate children nodes (orange) and their children nodes (green) together to form a unified prompt, and then input into BERT. Finally, we realize the prediction of various types of nodes through joint labeling.}
	\label{fig3}
\end{figure*}
\subsubsection{PPT: Taxonomy-path Prompt with Pre-trained model}

To address the limitations of the TEMP method, we proposed PPT (A Taxonomy Expansion Method Based on Taxonomy Path Prompt and Pre-Trained Model). Our approach includes a few improvements: 

Utilization of local information - Instead of using the entire taxonomy path, we use the local information nodes $l_{p}$ closest to the nodes. For example, in Figure \ref{fig2}, for the node ``Archeology", the local information nodes would be ``Archeology" and "Anthropology". When the depth of the taxonomy path is less than two, we take only one node. 
{
\begin{equation}
l_{p} = local(P) = \left \{ {N_{D-1},N_{D}} \right \} 
\end{equation}
 }

Improved pre-trained model utilization - We form a set of taxonomy path points $P_{Scocial Science} =$ (Archeology-Anthropology-Social Science) by combining the local information points of each node and the node Social Science to be extended. We then generate the appropriate natural language $S_{Gen}$ using a prompt function. 
{
\begin{equation}
S_{Gen}(q,l_{p}) = Prompt(q,l_{p})
\end{equation}
 }
where $q$ is the node to be expanded and $Prompt$ is a function to generate natural language from prompts.
For example, $S_{Gen}(q,l_{p}) =$ "Social Science including Anthropology, and Anthropology including Archeology". We feed this generated language $S_{Gen}$ into the pre-trained model, rank the results in the same way as TEMP, and use the highest score as the parent node of the given node. 

{
\begin{equation}
\operatorname{Encoder}(S_{Gen})=v^{'}_{[\mathrm{CLS}]}, v^{'}_{1}, \ldots, v^{'}_{w}
\label{eq2}
\end{equation}
 }
 The encoder results are as above, where $w$ is the number of output vectors.
 We trained the model with Margin Ranking Loss (MRL), which is defined as follows:

{\small
\begin{equation}
\mathcal{L}= \\
\sum_{P \in \mathcal{P}^{+}} \sum_{P^{\prime} \in \mathcal{P}^{-}} \max \left(0, f\left(P^{\prime}\right)-f(P)+\gamma\left(P, P^{\prime}\right)\right)
\end{equation}
 }
where $\mathcal{P}^{+}$ is the set of taxonomy-paths in the taxonomy, $\mathcal{P}^{-}$ is the set of negative samples, and $\gamma\left(P, P^{\prime}\right)$ is a function designed for the margin between positive and negative taxonomy-paths. To capture the semantic similarity of different taxonomy-paths, we follow TEMP to set a dynamic margin function based on the semantic similarity as follows:

\begin{equation}
\gamma\left(P, P^{\prime}\right)=\left(\frac{\left|P \cup P^{\prime}\right|}{\left|P \cap P^{\prime}\right|}-1\right) * k
\end{equation}
where k is a parameter used to adjust margins (usually between 0.1 and 1).
\begin{table*}[h]
	\centering \setlength{\tabcolsep}{1.5mm}
 
   \begin{tabular}{cc|ccc|ccc|ccc}
    \hline
    \multicolumn{2}{c|}{Dataset}                                                                                    & \multicolumn{3}{c|}{\textbf{Environment}}     & \multicolumn{3}{c|}{\textbf{Food}}            & \multicolumn{3}{c}{\textbf{Science}}          \\ \hline
    \multicolumn{2}{c|}{Metric}                                                                                     & Wu\&P         & MRR           & Acc           & Wu\&P         & MRR           & Acc           & Wu\&P         & MRR           & Acc           \\ \hline
    \multicolumn{1}{c|}{\multirow{6}{*}{\begin{tabular}[c]{@{}c@{}}Leaf \\ Nodes\end{tabular}}} & BERT+MLP          & 47.9          & 21.5          & 11.1          & 47.0          & 14.9          & 10.5          & 43.6          & 15.7          & 11.5          \\
    \multicolumn{1}{c|}{}                                                                       & TaxoExpan         & 54.8          & 32.3          & 11.1          & 54.2          & 40.5          & 27.6          & 57.6          & 44.8          & 27.8          \\
    \multicolumn{1}{c|}{}                                                                       & STEAM             & 69.6          & 46.9          & 36.1          & 67.0          & 43.4          & 34.2          & 68.2          & 48.3          & 36.5          \\
    \multicolumn{1}{c|}{}                                                                       & TMN               & 54.0          & 43.6          & 35.0          & 65.9          & 47.2          & 34.7          & 75.9          & 53.2          & 41.9          \\
    \multicolumn{1}{c|}{}                                                                       & TEMP-BERT         & 75.9          & 62.0          & 49.0          & 78.3          & 57.1          & 45.2          & 84.6          & 64.6          & 54.4          \\
    \multicolumn{1}{c|}{}                                                                       & ATTEMPT-PPT(ours) & \textbf{82.3} & \textbf{75.1} & \textbf{65.4} & \textbf{78.4} & \textbf{58.1} & \textbf{46.5} & \textbf{86.6} & \textbf{70.7} & \textbf{61.2} \\ \hline
    \multicolumn{1}{c|}{\multirow{2}{*}{\begin{tabular}[c]{@{}c@{}}All\\ Nodes\end{tabular}}}   & TEMP-BERT         & 64.8          & 33.5          & 15.6          & 84.3 & 41.1          & 37.4          & 88.0          & 52.9          & 43.4          \\
    \multicolumn{1}{c|}{}                                                                       & ATTEMPT-PPT(ours) & \textbf{81.1} & \textbf{52.5} & \textbf{37.5} & 84.3 & \textbf{42.6} & \textbf{39.0} & \textbf{90.9} & \textbf{60.0} & \textbf{52.8} \\ \hline
    \end{tabular}
    \caption{Baseline comparison on the three datasets in stage one (in \%)}
	\label{table1}
\end{table*}
 
\subsection{Stage two: Children Finding}
The second stage of ATTEMPT is to identify all the children nodes of a given node in the taxonomy. For example, finding the children nodes "Anthropology" and "Civics" for the node "Social Science", as shown in Figure \ref{fig2}. We propose two methods for this stage: PWC and MNL. 

\subsubsection{PWC: Pair-wise Classification}
In the second stage, we identify all the child nodes of a given term. To do this, we form pairs of possible ``hypernym-hyponym" term pairs from the node to be expanded (red node) and each candidate child node (orange node, child of the parent identified in the first stage). These term pairs are connected with the special token $[SEP]$ and fed into a pre-trained language model such as BERT. An example can be seen in Figure \ref{fig2}, where the node to be classified is ``Social Science" and the orange candidate child nodes are ``natural science," ``anthropology" and ``civics."

We use the pre-trained model to perform binary classification to determine whether the term pairs have a ``hypernym-hyponym" relationship or not. The traditional cross-entropy function is used as the loss function to train the classification model. This method is simple, because the pre-trained model has been trained on a large corpus already and it can identify whether the term pairs have a hierarchical relationship or not. This method is called Pair-wise classification.

\subsubsection{MNL: Multiple Nodes Labeling}
MNL is a new approach that addresses the problem of identifying multiple children of a given node in the taxonomy. There are two main challenges: determining whether a node has children and how many children it has, and identifying as many children as possible if there are multiple children.

To address these challenges, we first determine whether the given node is a leaf node (has no children) and if so, the second stage ends. If there are multiple children, we treat this as a multiple-choice problem and model it as a sequential labeling task. As shown in Figure \ref{fig3}, we extract the possible siblings, children, and grandchildren (orange and green nodes) of the given node to make use of local information. We then use a prompt function to convert these three types of nodes into natural language (e.g., ``Natural Science - Chemistry, Physics" is converted to ``Natural Science, and it including Chemistry and Physics").

We concatenate the node to be expanded ``Social Science" with all the sentences generated by the prompt, and then feed this into the pre-trained model. Since the model was trained on a large corpus of natural language, the input of natural language is consistent with the pre-training phase, which helps to fully utilize the hidden information of the model and correctly identify the contextual relationships. The addition of local information provides additional context to the model, which allows it to make more accurate predictions about the children of the given term.

\begin{table*}[h]
    \small
	\centering
    \begin{tabular}{cc|cc|cc|cc}
    \hline
    \multicolumn{2}{c|}{Dataset}                                                                                                                & \multicolumn{2}{c|}{\textbf{Environment}} & \multicolumn{2}{c|}{\textbf{Science}} & \multicolumn{2}{c}{\textbf{Food}} \\ \hline
    \multicolumn{2}{c|}{Metric}                                                                                                                 & acc                 & Avg(F1)               & acc               & Avg(F1)             & acc             & Avg(F1)           \\ \hline
    \multicolumn{1}{c|}{}                                                                            & PWC                                      & 21.5                & 50.8                & 20.0              & 54.3              & 12.5            & 32.1            \\
    \multicolumn{1}{c|}{\multirow{-2}{*}{\begin{tabular}[c]{@{}c@{}}Non-Leaf \\ Nodes\end{tabular}}} & {\color[HTML]{494949} ATTEMPT-MNL(ours)} & \textbf{46.2}       & \textbf{68.7}       & \textbf{33.3}     & \textbf{59.8}     & \textbf{37.5}   & \textbf{69.6}   \\ \hline
    \multicolumn{1}{c|}{}                                                                            & PWC                                      & 47.1                & 56.3                & 58.1              & \textbf{74.65}    & 60.3            & 74.8            \\
    \multicolumn{1}{c|}{\multirow{-2}{*}{\begin{tabular}[c]{@{}c@{}}All\\ Nodes\end{tabular}}}       & {\color[HTML]{494949} ATTEMPT-MNL(ours)} & \textbf{64.7}       & \textbf{82.8}       & \textbf{61.3}     & 74.09             & \textbf{73.3}   & \textbf{84.5}   \\ \hline
    \end{tabular}
    \caption{Baseline comparison on the three datasets in stage two (in \%)}
	\label{table2}
\end{table*}

\begin{table*}[h]
	\centering
    \begin{tabular}{c|cc|cc|cc}
    \hline
    Dataset       & \multicolumn{2}{c|}{\textbf{Environment}} & \multicolumn{2}{c|}{\textbf{Science}} & \multicolumn{2}{c}{\textbf{Food}} \\ \hline
    Metric        & acc                 & Avg(F1)               & acc               & Avg(F1)             & acc              & Avg(F1)           \\ \hline
    Baseline      & 11.8                & 14.7                & 22.5              & 23.3              & 25.9             & 30.9            \\ \hline
    ATTEMPT(ours) & \textbf{12.0}       & \textbf{15.0}       & \textbf{22.6}     & \textbf{28.1}     & \textbf{26.7}    & \textbf{32.9}   \\ \hline
    \end{tabular}
    \caption{Comparison of the baseline method and ATTEMPT in the overall process(in \%)}
	\label{table3}
			\vspace{-0.3cm}
\end{table*}

\section{Experiments}
In this section, we first describe the experimental setup and implementation details in Section \ref{4.1} and Section \ref{4.2}. We then present the results of our experiments in Section \ref{4.3}, including a comparison of our approach to the baseline method. To further understand the contribution of different components of our approach, we conduct ablation experiments in Section \ref{4.4} to investigate the effectiveness of using local information and prompts in ATTEMPT.

\subsection{Experimental Setup}
\label{4.1}
\textbf{Datasets.} We conducted experiments on two datasets that include six domains and two types of nodes. The first dataset is the Semeval-2016 task 13 dataset, which was used to evaluate the performance of expanding leaf nodes in stage one. We compared our method to previous approaches such as TEMP \cite{liu-etal-2021-temp} and STEAM \cite{10.1145/3394486.3403145}, which have also been tested on this dataset for leaf node expansion.

To evaluate the expansion of non-leaf nodes, we constructed a new dataset based on Semeval, as there are limited previous datasets that are relevant to this task. This dataset was specifically designed for the purpose of non-leaf node expansion and evaluation.

The following is a description of the two datasets: 1)
We used the dataset from Semeval-2016 task 13 \footnote{https://alt.qcri.org/semeval2016/task13/}, which contains three English datasets for the environment, science, and food domains. We followed the setup as in \cite{10.1145/3394486.3403145} and used the randomly-grown taxonomies for self-supervised learning, and sampled 20\% of the leaf nodes for testing. We used this dataset to compare our method with other taxonomy extension methods for leaf nodes. 2) As there is limited data available for non-leaf node expansion, we reconstructed the original data. We defined nodes with one parent and no children as leaf nodes and nodes with one parent and at least one child as non-leaf nodes. More details about the dataset are provided in Appendix \ref{dataset}.

\noindent\textbf{Metrics.} For the parent finding process in stage 1, we followed the evaluation strategy of \cite{10.1145/3394486.3403145} using Accuracy, Mean reciprocal rank (MRR), and Wu \& Palmer similarity (Wu\&P) to evaluate our methods. Accuracy (ACC) measures the count of parent or child nodes that are accurately predicted.
MRR calculates the average of reciprocal ranks of the true taxonomy path.
Wu\&P measures the semantic similarity between the predicted taxonomy path and the truth taxonomy-path.

For stage two, we proposed two metrics for evaluating the effectiveness of this phase. One is ACC, which represents whether all children can be found or not. The second one is Avg F1, which can further evaluate how many children are found for a given node.
$\operatorname{Avg} (\mathrm{F} 1)=\frac{1}{n} \sum_{i=1}^{n} \mathrm{~F} 1 $
  
\noindent\textbf{Compared Methods.} We compare with the following methods:

$\bullet$ \textbf{BERT+MLP} The method extracts terms embeddings from BERT and then feeds them into a multilayer perceptron (MLP) to predict their relationship. 

$\bullet$ \textbf{TEMP} \cite{liu-etal-2021-temp} One state-of-the-art taxonomy expansion framework which predicts new concepts' position by ranking the generated taxonomy paths. The first method that employs pre-trained contextual encoders in taxonomy construction and hypernym detection problems.

$\bullet$ \textbf{STEAM} \cite{10.1145/3394486.3403145} A taxonomy expansion framework that leverages natural supervision in the existing taxonomy for expansion. 

$\bullet$\textbf{TaxoExpan} \cite{10.1145/3366423.3380132} A self-supervised method for encoding local structures in seed taxonomy using location-enhanced graph neural networks. 

$\bullet$ \textbf{TMN} \cite{zhang2021tmn} A Triplet Matching Network (TMN) that finds suitable hypernym, hyponym word pairs for a given query concept. 
\begin{table*}[h]
	\centering \setlength{\tabcolsep}{1.5mm}
    \begin{tabular}{c|ccc|ccc|ccc}
    \hline
    Dataset       & \multicolumn{3}{c|}{\textbf{Environment}}                   & \multicolumn{3}{c|}{\textbf{Science}}         & \multicolumn{3}{c}{\textbf{Food}}             \\ \hline
    Metric        & Wu\&P                       & MRR           & Acc           & Wu\&P         & MRR           & Acc           & Wu\&P         & MRR           & Acc           \\ \hline
    no path nodes & 81.8                        & 73.2          & 59.6          & 81.0          & 67.2          & 56.5          & 78.5          & 57.5          & 44.1          \\ \hline
    no prompt     & {\color[HTML]{494949} 79.0} & 64.8          & 51.9          & 84.1          & 67.0          & 56.5          & 78.4          & 56.9          & 45.1          \\ \hline
    ATTEMPT-PPT(ours)     & \textbf{82.3}               & \textbf{75.1} & \textbf{65.4} & \textbf{86.6} & \textbf{70.7} & \textbf{61.2} & \textbf{78.7} & \textbf{58.1} & \textbf{46.5} \\ \hline
    \end{tabular}
    \caption{Results of ablation experiments on the three dataset in stage one (in \%)}
	\label{table4}
\end{table*}

\begin{table*}[h]
	\centering \setlength{\tabcolsep}{1.5mm}
    \begin{tabular}{c|cc|cc|cc}
    \hline
    Dataset                              & \multicolumn{2}{c|}{\textbf{Environment}} & \multicolumn{2}{c|}{\textbf{Science}} & \multicolumn{2}{c}{\textbf{Food}} \\ \hline
    {\color[HTML]{494949} Metric}        & acc                 & Avg(F1)               & acc               & Avg(F1)             & acc             & Avg(F1)           \\ \hline
    no prompt                            & 52.9                & 53.7                & \textbf{62.0}     
    & \textbf{75.0}     & 52.7            & 63.4            \\
    {\color[HTML]{494949} no grandchild} & 52.9                & 52.9                & 51.6              & 62.3              & 51.1            & 59.8            \\ \hline
    ATTEMPT-MNL(ours)                       & \textbf{64.7}       & \textbf{82.8}       & 61.3              & 74.2             & \textbf{73.3}   & \textbf{84.5}   \\ \hline
    \end{tabular}
    \caption{Results of ablation experiments on the three dataset in stage two (in \%)}
	\label{table5}
			\vspace{-0.3cm}
\end{table*}
\subsection{Implementation Details}
\label{4.2}
We present the PPT method for the first stage of leaf node expansion, which is based on TEMP (TEMPs' code link \footnote{https://github.com/liu-zichen/TEMP}). We use BERT (bert-base-uncased) as the pre-trained language model and split the terms into 10\% for validation and 10\% for testing. To expand the full type of nodes, both leaf and non-leaf, we use the new data introduced previously and select the same number of leaf and non-leaf nodes as the test set. We use the default optimal hyperparameters of the original TEMP authors and experiment with different learning rates to obtain the best performance. We also use multiple prompts (see Appendix \ref{Prompts}) according to the settings of Musubu ~\cite{takeoka-etal-2021-low}, and take the average result as the experimental result. To reduce the impact of randomness, we repeat the experiment three times. 

For the MNL method in stage two, we connect the nodes to be expanded (red), the candidate child nodes (orange), and the child nodes of the candidate nodes (green) and generate natural language by way of prompt. The generated natural language is fed into the pre-training model and labelled. We label the real children of a given node as 1, the sibling nodes as 0, and ignore the computational loss for all the rest of the nodes. In addition, if a term has multiple tokens and one of the tokens is marked as one by the model, we mark all those tokens as child nodes. See Appendix \ref{id} for more details.

\subsection{Experimental Results}
\label{4.3}
As shown in Table \ref{table1}, our method PPT outperforms the existing TEMP model significantly on both leaf and non-leaf nodes. For leaf nodes, we improved the TEMP model by 8.2\%, 6.7\%, and 2.8\% on Acc, MRR, and Wu\&P, respectively. For all types of nodes, the improvement is 11.0\%, 9.2\%, and 6.4\%, respectively.

The comparison results of the two methods tested in the child discovery phase are presented in Table \ref{table2}. For leaf nodes, the MNL method improves Acc and Avg(F1) by 21\% and 20.3\%, respectively, compared to the pair-wise classification method over the three benchmark datasets. For all type nodes, the improvement is 11.3\% and 11.9\%, respectively.

Table \ref{table3} presents the comparison results between the baseline method and our ATTEMPT. The baseline method achieves 14.7\%, 23.3\%, and 30.9\% in Avg(F1) metrics for the three datasets of environment, science, and food, respectively. Our ATTEMPT method improved the Avg(F1) by an average of 2.4\% over the baseline. The low results in Table 3 are due to the challenging nature of the task. To obtain the correct parent node, all child nodes must be successfully identified. This highlights the potential for further improvement.

\subsection{Ablation Studies}
\label{4.4}
To verify local information and prompt effectiveness, we compare and test the changes in experimental results with/without these two types of information on both stages.

\textbf{Local Information} As shown in Table \ref{table4}, after removing the path nodes, the PPT method in stage 1 decreases on average by 4.3\%, 2\%, and 2.1\% on acc, mrr,wu\&p, respectively, on the three datasets. Table \ref{table5} also shows that the MNL method decreases by 14.6\% and 22.1\% on average on accuracy and average F1 score, respectively, after removing the grandchild node information in the child finding stage. We found that local information is essential in both the first and second phases, particularly in the second child lookup phase. Removing local information brings about a significant performance degradation, which may be attributed to our method's modelling of relationships. The individual nodes are closely associated in our MNL method.

\textbf{Prompt} In Table \ref{table4}, the PPT method with prompt removal decreased in acc, mrr, wu\&p by 6.5\%, 5.1\%, and 2.0\% on average, respectively. Meanwhile, in the second stage, the MNL method decreased 8.6\% and 15.6\% for accuracy and average F1 metrics, respectively, after prompt removal. The scientific data in the second stage showed a slight performance improvement after prompt removal, which we speculate may be due to insufficient data and pre-trained corpus. Overall, the prompt is essential for the parent finding process in the first stage and the child finding process in the second stage.

\section{Conclusion}
This paper proposes a two-stage taxonomy completion method based on pre-trained Language models (ATTEMPT), which effectively inserts the new concept in the correct position by finding a parent node then labeling children nodes. In addition, we use prompt to generate natural language information suitable for the pre-trained model further to improve the effectiveness of parent node recognition and children labeling for the given node. Our experiments on two types and three domains with six datasets show that our method can enhance the effectiveness of locating the position of a given node in existing taxonomies. Furthermore, the efficacy of local information and prompts in ATTEMPT is also demonstrated by ablation experiments. In conclusion, our proposed ATTEMPT method is an effective approach for taxonomy completion, and it can be further improved with more comprehensive datasets.

\section*{Limitations}
Since ATTEMPT uses the pre-trained language model to complete the taxonomy, the expansion effect is limited by the model. Generally, pre-trained models with more knowledge scales are better (e.g., BERT-Large V.S. BERT-Base-uncased). However, our paper focuses on how to fully use the knowledge of the pre-trained model rather than verifying whether more knowledge scales better or not. Based on the above, this paper does not conduct more related research (in fact, TEMP \cite{liu-etal-2021-temp} has been compared and reached similar conclusions). In addition, the selection of prompts will also affect the expansion effect. For the convenience of comparison, we have selected several basic prompts (the same as Musubu~\cite{takeoka-etal-2021-low}) for experimentation. In future work, we plan to study how to construct or select better prompts for classification expansion. We do not consider the situation of multi-parent nodes according to the TEMP \cite{liu-etal-2021-temp} settings. And according to our statistics, there are only a few multi-parent nodes in the Semeval-2016(task-13) datasets (1/3843). We will continue investigating how to make better use of the pre-trained model knowledge to solve the taxonomy completion problem.

\bibliography{anthology,custom}
\bibliographystyle{acl_natbib}

\appendix
\vspace{-0.4cm}
\section{Dataset}
\vspace{-0.2cm}
\label{dataset}

The original dataset and our reconstructed dataset statistics are in Table \ref{table6} and \ref{table7}. 

\begin{table}[h]
\small
\begin{tabular}{crrr}
\hline
Dataset         & Environment & Science & food    \\ \hline
$|\mathcal{N}|$             & 261         & 429     & 1486    \\ 
$|\mathcal{L}|$             & 200         & 306     & 1161    \\
$|\mathcal{D}|$            & 6           & 8       & 8       \\
$|\Delta_{leaf}|$       & 3.79        & 5.17    & 5.37    \\
$|\Delta_{non}|$       & 2.97        & 4.41    & 4.78    \\
train(leaf/non) & 148/59      & 226/112 & 893/284 \\
test(leaf/non)  & 52/0        & 85/0    & 297/0   \\ \hline
\end{tabular}
\caption{Statistics of the original taxonomy datasets for evaluation. $|\mathcal{N}|$ and $|\mathcal{L}|$ are the number of nodes and leaf nodes in the taxonomy. $|\mathcal{D}|$, $|\Delta_{leaf}|$ and $|\Delta_{non}|$ indicate the depth of the taxonomy and the average depth of leaf nodes and non-leaf nodes respectively.Train(leaf/non) and test(leaf/non) represents the proportion of leaf and non-leaf nodes in train data and test data.}
	\label{table6}
   \vspace{-0.6cm}
\end{table}

\begin{table}[h]
\small
\begin{tabular}{crrr}
\hline
Dataset         & Environment & Science & food     \\ \hline
$|\mathcal{N}|$             & 261         & 429     & 1486     \\
$|\mathcal{L}|$             & 200         & 306     & 1161     \\
$|\mathcal{D}|$             & 6           & 8       & 8        \\
$|\Delta_{leaf}|$       & 3.79        & 5.17    & 5.37     \\
$|\Delta_{non}|$        & 2.96        & 4.41    & 4.05     \\
train(leaf/non) & 187/46      & 292/94  & 1124/214 \\
test(leaf/non)  & 13/13       & 21/21   & 74/74    \\ \hline
\end{tabular}
\caption{Statistics of the new taxonomy datasets for evaluation. }
	\label{table7}
  \vspace{-0.6cm}
\end{table}

To prevent the test data from being leaked during training and to thoroughly test the generalization ability of the model when encountering unseen data, we split each original taxonomy tree into two subtrees, one for training and one for testing. For example, the left subtree of the scientific taxonomy in Figure \ref{fig2}, natural science and its children, is used as the test subtree, and the rest is used for training. Specifically, we select the subtree with 20\% of the number of nodes of the current taxonomy tree as the subtree for testing and ignore too many leaf nodes to ensure the ratio of leaf nodes to non-leaf nodes is 1:1. Too many leaf nodes will make the child finding stage degenerate into an expansion of leaf nodes, and the model will be easily overfitting. And too few leaf nodes will make the test inadequate, so we use equal leaf and non-leaf node data as the test.

In the training and testing phases, we dig out the node to be expanded in the current taxonomy tree, and if the node has N children, these N children are reassigned to the original parent of the node to be expanded as child nodes. We ignore the case of double parent nodes because their existence is too rare. Only one node in the three datasets containing more than 2000 nodes in our experiments has a dual-parent node. We will consider this case further in our future work.

\section{Implementation Details}
\label{id}
For the fairness of the experiment, we follow the setting of TEMP \cite{liu-etal-2021-temp}. We use 10\% terms for validating and 10\% for testing. For each benchmark, we try various learning rates and report the best performance. We use multiple prompts to experiment and select the average result as the experimental result. We repeated the experiment three times to reduce the impact of randomness. We train the model using the Pytorch \footnote{\url{https://pytorch.org}} \cite{NEURIPS2019_bdbca288} on the NVIDIA RTX3090 GPU. For all methods, the bert-base-uncased \footnote{\url{https://huggingface.co/bert-base-uncased}} model are chosen for feature extraction. The pretrained contextual encoders are of base size with 12 layers. We use the AdamW \cite{IlyaLoshchilov2018DecoupledWD} as the optimizer with the warm-up \cite{7780459}, and fine tune the whole model with a learning rate of 2e-5. The dropout \cite{NitishSrivastava2014DropoutAS} of 0.1 is applied to prevent overfitting.
\section{Prompt Details}
\label{Prompts}

\vspace{-0.4cm}
\begin{table}[h]
\centering \small
\begin{tabular}{cc}
\hline
Name         & Prompt     \\ \hline
Such-as             & Y such as X and Z     \\
One-of             & X is one of Y, and Z is one of  Y      \\
Especiaally & Y, especially X and Z\\
Is-a & X is a Y, and Z is a Y\\
Including & Y including X and Z\\
\hline
\end{tabular}
\caption{List of prompts used in the experiments. Y denotes a parent term of a term X and Z }
	\label{table8}
	\vspace{-0.4cm}
\end{table}

\end{document}